\documentclass{article}

\usepackage{arxiv}

\usepackage[utf8]{inputenc} 
\usepackage[T1]{fontenc}    
\usepackage{hyperref}       
\usepackage{url}            
\usepackage{booktabs}       
\usepackage{amsfonts}       
\usepackage{nicefrac}       
\usepackage{microtype}      
\usepackage{lipsum}
\usepackage{graphicx}
\usepackage{siunitx}
\usepackage{bm}

\title{Exploring environment exploitation for self-reconfiguration in modular robotics}

\author{
Philippe Martin Wyder \\
Department of Applied Mathematics \\
University of Washington \\
Seattle, WA, USA \\
\texttt{philippe.wyder@columbia.edu} \\
\And
Haorui Li \\
Department of Mechanical Engineering and Applied Mechanics \\
University of Pennsylvania \\
Philadelphia, PA, USA \\
\texttt{haoruil@seas.upenn.edu} \\
\And
Andrew Bae \\
Department of Mechanical Engineering \\
University of Nevada, Las Vegas \\
Las Vegas, NV, USA \\
\texttt{andrew.bae@unlv.edu} \\
\And
Henry Zhao \\
Department of Mechanical Engineering and Applied Mechanics \\
University of Pennsylvania \\
Philadelphia, PA, USA \\
\texttt{henryzhao@seas.upenn.edu} \\
\And
Mark Yim \\
Department of Mechanical Engineering and Applied Mechanics \\
University of Pennsylvania \\
Philadelphia, PA, USA \\
\texttt{yim@seas.upenn.edu} \\
}

\begin{document}
\maketitle
\begin{abstract}
Modular robotics research has long been preoccupied with perfecting the modules themselves---their actuation methods, connectors, controls, communication, and fabrication. This inward focus results, in part, from the complexity of the task and largely confines modular robots to sterile laboratory settings. The latest generation of truss modular robots, such as the Variable Topology Truss and the Truss Link, have begun to focus outward and reveal a key insight: the environment is not just a backdrop; it is a tool. In this work, we shift the paradigm from building better robots to building better robot environment interactions for modular truss robots. We study how modular robots can effectively exploit their surroundings to achieve faster locomotion, adaptive self-reconfiguration, and complex three-dimensional assembly from simple two-dimensional robot assemblies. By using environment features---ledges, gaps, and slopes---we show how the environment can extend the robots' capabilities. Nature has long mastered this principle: organisms not only adapt, but exploit their environments to their advantage. Robots must learn to do the same. This study is a step towards modular robotic systems that transcend their limitations by exploiting environmental features.
\end{abstract}

\keywords{Modular Robot, Reconfiguration, Truss Robot}

\section{Introduction}

\begin{figure*}[t]
    \centering
    \includegraphics[width=6.6in]{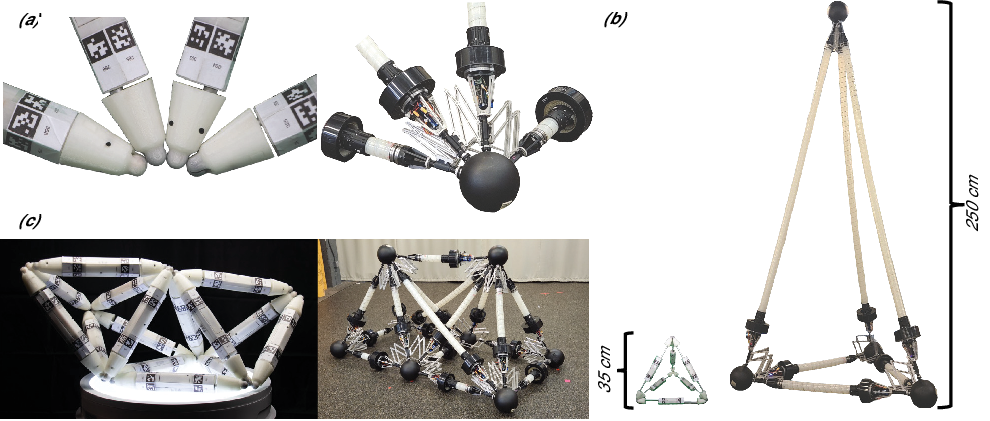}
    \caption{Comparison between the Truss Link and the VTT platform. (a) The magnetic connectors of the Truss Link (left) and the mechanical connector of the VTT (right). (b) Size comparison between a fully expanded Truss Link tetrahedron robot (left) and a VTT robot with its top links fully expanded (right). (c) Example images of complex, manually assembled topologies: a robot comprising fifteen Truss Links (left) and an eighteen-module VTT robot (right). }
    \label{Fig_VTT_TrussLink_Comparison}
\end{figure*}

The field of robotics takes inspiration from nature: robot dogs, humanoid robots, reinforcement learning, robot swarms, and modular robots.
While biological creatures have evolved with and in response to their environment, robots are traditionally purpose-built for specific environmental conditions. 
Self-reconfiguring and self-assembling Modular robots are an exception, they are able to adapt their morphology and can even grow by absorbing more material \cite{Wyder2025RobotMetabolism} 
or adapt their body to cope with their environment \cite{liu2023smores}.
In this study, we find that the reverse, adapting environmental conditions to an existing robot and thereby identifying synergistic relationships, is largely overlooked in robotics research. 
We can unlock new robot capabilities by identifying environmental conditions that a robot can recognize and then exploit.
In particular, we focus on truss modular robots that can exploit their environment to self-reconfigure or self-assemble.

Robots that can exploit environmental features are not new.
There are notable examples of robots that can exploit environmental features to their advantage, such as robots with micro spine grippers that can scale cliffs and vertical walls \cite{spenko2008biologically, nadan2024loris}, and soft and compliant robots that guide their movement on walls and obstacles \cite{batra2024vibrating, li2019particle, lee2017soft, bao2018soft, walker2020soft, bilodeau2017self}. 
However, most robots are commonly developed to ``cope'' with their environment rather than exploit it. 
In robotics research, overcoming obstacles by squeezing under low passages or through narrow gaps are commonly studied environment interactions \cite{hoeller2024anymal, sihite2023multi, bucki2019design, falanga2018foldable}. 
We believe that the study of exploitable environment features bears the potential to unlock new breakthroughs, even in modular robotics and robotics more generally.

Animals naturally evolve the ability to  exploit favorable environmental conditions.
Beavers tend to place their dams in narrowing sections of a river or stream with slower water flow \cite{bashinskiy2020beavers, larsen2021dam}. 
Similarly, spiders build webs in areas that support their web type—for example, between branches and places in the paths of prevailing winds to increase their chances of capturing prey \cite{eberhard2020spider}. 
Just as animals evolve to exploit their environment for survival, robots can do the same to achieve their objectives---for example, self-reconfiguration.

Modular robots have the potential to exploit environments better than conventional robot platforms due to their flexibility \cite{seo2019modular},
\cite{yim2002connecting}.
However, previous modular robots self-reconfigure to cope with the complexity of their environment, but they did not exploit it \cite{murata2002m, murata2001self, liu2023smores, davey2012emulating}.

One example modular robot system is the Variable Topology Truss (VTT) system  \cite{spinos2021topological}.
The concept of the VTT system includes self-reconfiguration feature to choose optimal topology for various tasks.
Our prior work considered VTT topology reconfiguration in 3-dimensions (3D) \cite{park2020reconfiguration, spinos2021topological, yoon2024compliant}. 
Having two modules dock in 3D space has been shown, but has been difficult to achieve reliably \cite{roufas2001six}.
This paper studies self-reconfiguration for a truss-style modular robotic systems, with the goal of reconfiguring robots from 2-dimensional (2D) configurations into a 3D configuration.
Unlike in 3D reconfiguration, the robot doesn't need to support its own 3D structure while rearranging truss attachments. 
3D space reconfiguration in general requires control in a six-dimensional space, and due to low stiffness often has to deal with dynamic vibration of cantilever beams, while 2D reconfiguration can be quasi-static and only requires control over three dimensions.
Therefore, conducting a reconfiguration in 2D eases the constraints and greatly reducing the complexity of the reconfiguration task.

This work focuses on two robotic platforms that can exploit their environment to self-assemble or self-reconfigure to form topologies that would otherwise be unlikely or difficult to achieve. We are exploring how the design space of new capabilities and new configurations is expanded as a result of a suitable environmental feature. The demonstrations from this paper highlight the transformative potential effective environment exploitation can have on a modular robot's capabilities and design transformations.

First, we introduce the Truss Link and the VTT robot platforms (see Fig. \ref{Fig_VTT_TrussLink_Comparison}). 
Next, we introduce the environments and features used in our experiments, as well as the simulation environments used in our experiments. 
We present two examples of how a truss robot exploits environmental features to transform a topology that is the result of 2D reconfiguration or assembly to achieve a 3D structure. 
Finally, we examine environment exploitation on the example of a robot assisting another in reconfiguring from a 2D topology to a 3D topology.

This work provides both theoretical and conceptual contributions to the field of modular robotics, but also improvements in hardware and demonstrated robot capability. 
We present an enhanced spherical joint linkage on the 2D VTT to enable the automatic attachment and detachment from other modules. 
Further, we share two unseen demonstrations of robot self-reconfiguration: planar VTT docking and 2D to 3D self-reconfiguration using the mound environment.

\section{Materials and methods} 

First, we compare the two robot platforms used in our experiments: the Variable Topology Truss (VTT) \cite{yoon2024compliant, park2020reconfiguration, spinos2021topological} and the Truss Link  \cite{Wyder2025RobotMetabolism}. 
Next, we outline the experiments and their respective environment topologies.
As the focus is not on autonomy, the robots in all experiments were controlled using a combination of teleoperation and open-loop scripts. 
Developing automatic control methods for the demonstrations in this paper is an active area of research and is left for future work.

\subsection{Robot Platforms}

Both VTT and Truss Link are modular, self-reconfigurable truss robot systems that can form tetrahedral-style structures (see Fig. \ref{Fig_VTT_TrussLink_Comparison}).
VTT was developed with the goal of aiding in search and rescue operations by supporting compromised structures, and is able to operate on a wide range of spatial scales due to its linear spiral zipper actuator capable of expanding more than 14 times its own length.

\begin{table}[h]
\small\sf\centering
\caption{Comparison of VTT and Truss Links.\label{tbl:VTT_TrussLink_Comparison}}
\begin{tabular}{lcc}
\toprule
Feature & VTT & Truss Links \\
\midrule
Weight & \SI{2.8}{kg} & \SI{280}{g} \\
Min. Length &  \SI{94}{cm} & \SI{28}{cm} \\
Max. Length &  \SI{213}{cm} & \SI{48}{cm} \\
Expansion Actuators & 1 & 2 \\
Actuator Type & Spiral Zipper & Linear Servo \\
Max. Actuation Force & \SI{220}{N} & \SI{80}{N} \\
Symmetrical Design & No & yes \\
Connector Actuators & 2 & 0 \\
Connector Mechanism & Active & Passive \\
Connector Type & Mechanical & Magnetic \\
Connector sep. Force & N/A & \SI{13.6}{N} \\
Expansion Ratio & 227\% & 53\% \\
Battery Capacity & 1600mAh & 300mAh \\
Battery Runtime & 50-min & 40-min \\
Controller & ESP32-PICO D4 & Particle Photon \\
Connectivity & WIFI & WIFI \\
\bottomrule
\end{tabular}
\end{table}

In contrast, the Truss Link was developed as a research platform to study physical robot development and can expand 53\% in length---from \SI{28}{cm} to \SI{48}{cm}. 
See Table \ref{tbl:VTT_TrussLink_Comparison} for a side-by-side comparison of VTT and Truss Link specifications.

For docking during self-reconfiguration,
the VTT relies on a spherical linkage connector that requires careful alignment to form connections, while the Truss Link uses a free-form magnetic attachment mechanism that greedily attaches to any connector within close proximity. As a result, the VTT forms stronger structures than the Truss Link, but lacks the ease of self-assembly of Truss Links. Refer to \cite{yoon2024compliant, park2020reconfiguration, spinos2021topological} for further information on VTT and to \cite{wyder2024robotlinks} for an introduction to the Truss Link (formerly known as Robot Link).

In ideal truss structures, truss members support tensile and compressive loads, while the nodes support no moments. The magnetic connectors of the Truss Links act as free-moving spherical joints at the connection points and thus naturally enforce this. In the VTT robot, a spherical joint linkage performs the same function, allowing truss-members to move around the sphere while maintaining continuous contact. 

In both VTT and Truss Link reconfiguration can occur by having small groups of modules attach to each other to form larger robots or detach to reconfigure. These  robots are capable of independent locomotion and move around the environment to then meet and dock with other groups of modules.

\subsection{Planar Locomotion}
\begin{figure*}
    \centering
    \includegraphics[width=6.1in]{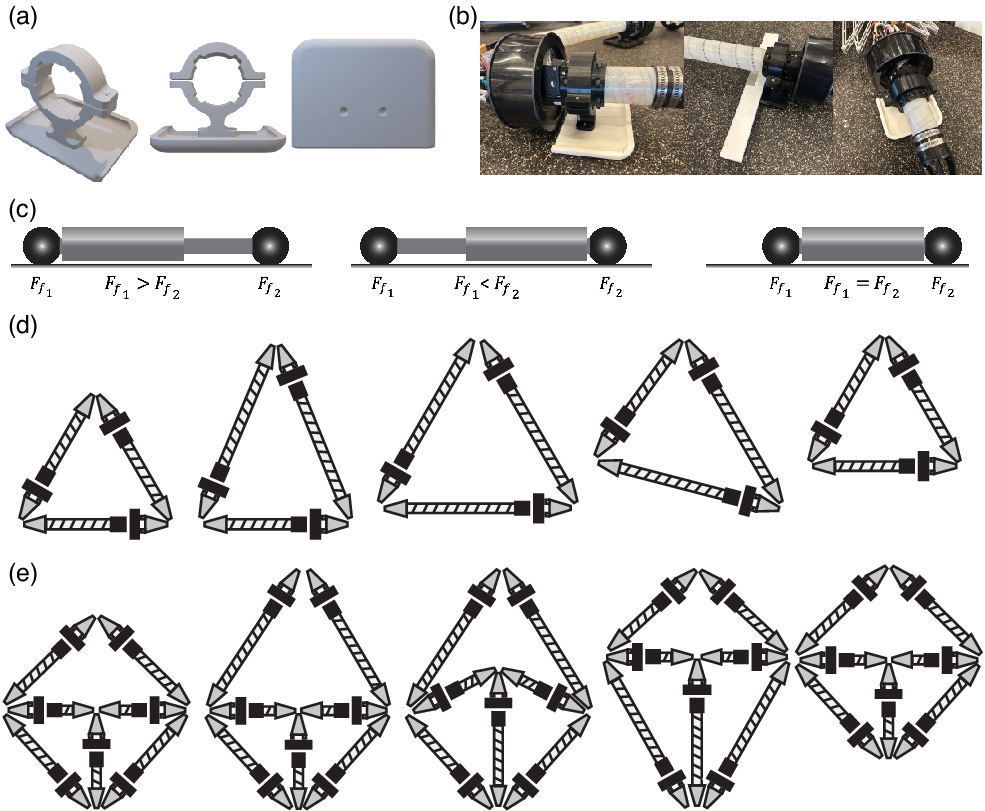}
    \caption{Crawling locomotion strategy. {\bf(a)} For VTT, sliding feet prevent the linear zipper actuator from touching the ground and provide an even sliding surface that contacts the ground during crawling motion. {\bf(b)} Sliding feet mounted on VTT. {\bf(c)} Truss Link inch-worm style crawling. 1.~Expands the front servo,  center of mass leans left. 2.~Contract the front servo while expanding the rear servo, center of mass leans right. 3.~Contract the rear servo ready for the next crawl cycle. $F_{f_1}$ and $F_{f_2}$ correspond to the friction force. {\bf(d)} VTT crawling locomotion in a triangular configuration: 1.~Initial configuration;  2.~Expanding one node towards the preferred moving direction (up) by extending the two top members;  3.~Expanding the member at the opposite side;  4.~Retracting one side;  5.~Retracting the other two members to return to the initial configuration. {\bf(e)} VTT crawling locomotion in square configuration; 1.~Initial configuration; 2.~Expanding one node up; 3.~Relocating the center node up; 4.~Moving the side nodes up; 5.~Pulling the rear node and returning to the initial configuration.}
    \label{Fig_VTTCreepLocomotion}
\end{figure*}
Both systems can crawl on flat terrain by repeatedly expanding and contracting their prismatic joints in specific sequences. In order to achieve forward progress, these robots exploit differential friction between contact points. If a robot were to symmetrically grow and shrink, there would be no progressive motion, but by expanding some parts, but not others in a group, the robots break frictional symmetry.

Unlike the VTT, Truss Links are actuated by two independent linear servos and are therefore capable of microlocomotion. This means that an individual Truss Link module can crawl forward and backwards, but not steer actively. The servos of a single module can execute a sequence of asymmetrical expand and contract motions to produce an inch-worm style crawl gate as illustrated in Fig. ~\ref{Fig_VTTCreepLocomotion}~(c). 

In 2D configurations and 3D configurations such as triangle or tetrahedron topologies, VTT and Truss Link robots can crawl on a 2D plane.
Assuming Coulomb friction on a uniform surface (equal friction coefficients at all contact points), the frictional forces are determined by the normal forces that result from the mass distribution.
The contact points supporting a higher share of the robot's mass experience higher maximum static friction.
Therefore, we can control which contact point will slide while the others remain stationary, by shifting the mass of the robot over the contact points that we want to keep stationary. The crawling locomotion strategy of the triangle and square configuration of the VTT is shown in Fig.~\ref{Fig_VTTCreepLocomotion}(d) and Fig.~\ref{Fig_VTTCreepLocomotion}(e) respectively.

\begin{figure*}
    \centering
    \includegraphics[width=6.1in]{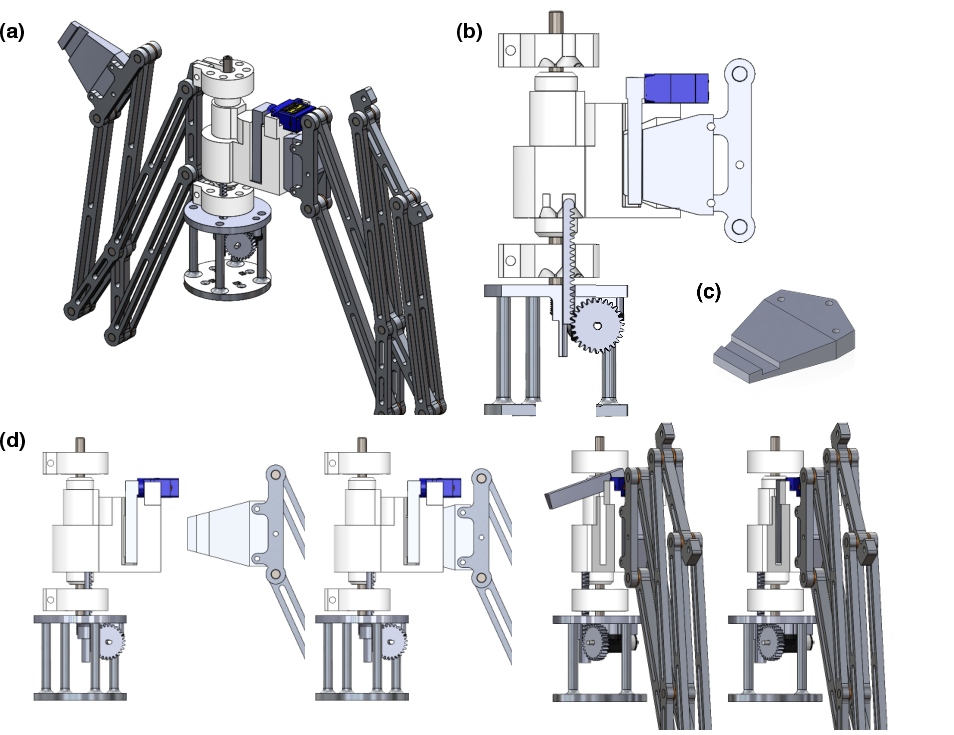}
    \caption{VTT docking mechanism design; (a) The VTT member-end with docking mechanism; (b) Rack and pinion mechanism for locking member rotation during docking; (c) Docking insert design; (d) Docking process between VTT member modules. Put the docking insert into the docking port by crawling locomotion (first and second figure); Close the locking lever to fix the docking part (third and fourth figure).}
    \label{Fig_VTTDocking}
\end{figure*}

3D Truss Link and VTT robots are also capable of other locomotion methods. When the VTT system is in 3D configurations, such as a tetrahedron or octahedron, it can use a rolling/tumbling locomotion \cite{park2019optimization, park2020polygon}. In a prior work, Wyder et al.~demonstrated tetrahedron tumbling as a means to overcome obstacles and ratchet tetrahedron motion on sloped surfaces as a faster mode of locomotion \cite{Wyder2025RobotMetabolism}. In this work, we re-analyze the tetrahedron formation experiments shown by Wyder et al. through the lens of environment exploitation and consider them in context of truss modular robot reconfiguration more generally.

To tumble, Truss Link and VTT robots change their shape such that the center of mass moves to be outside the support polygon, so the shape rolls over to the adjacent face. When tumbling Platonic solids such as tetrahedrons, octahedrons, etc. the impact from toppling decreases with the refinement level of the polyhedron. For example, the VTT octahedral configuration can significantly minimize its falling impact of the tumble compared to a tetrahedron.

\subsubsection{Truss Link Connections.}

The Truss Link uses permanent-magnet, free-form style connectors. 
Each connector contains a 1/2-inch-diameter Neodymium magnet sphere that can be passively retracted within the connector shell to ``deactivate`` the attraction force, when the Truss Link fully contracts the servo attached to the connector. The connector shell contains a conical reset spring that ensures that the magnet sphere is in the tip of the connector whenever the servo isn't fully retracted.
The magnet sphere is position constrained, but free to rotate in place, and therefore can move to an equilibrium position even when more than two connectors connect at one point. Truss Link connectors act as passive spherical joints; thus, a vertex requires three connections to be position constrained in 3D space. Due to the high-friction plastic combined with the strong magnet attraction force, connectors do not tend to slide easily along other connectors once connected.
\begin{figure*}
    \centering
    \includegraphics{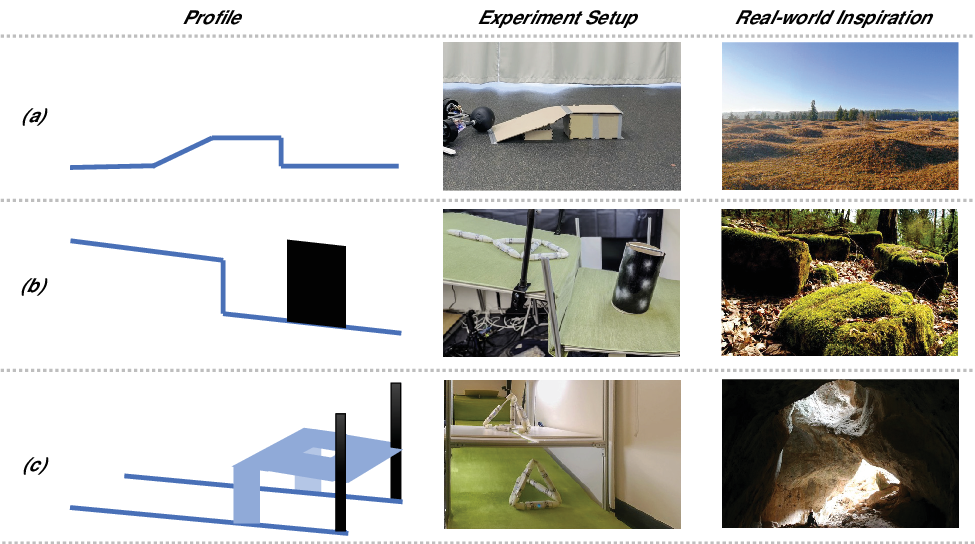}
    \caption{Experiment environments: (a) mound environment; (b) step environment; (c) skylight environment. The left column shows an abstraction of the experiment environment; the center column shows a picture of the physical environment used for the experiment, and the right-hand column shows a picture of a naturally occurring environment that inspired the experiment topology.}
    \label{Fig_RealVsArtificialEnv}
\end{figure*}
\subsubsection{VTT Connections.}

Fig.~\ref{Fig_VTTDocking} introduces the design and process of docking for the VTT system. The VTT's mechanical docking mechanism is actuated by two servos. One for locking the member's rotational joint, so the connector can align with its counterpart, and one for closing the connector latch and securing the connection.  The rotation of the members can be locked and unlocked using a rack-and-pinion mechanism (Fig.~\ref{Fig_VTTDocking}~(b)). After locking the rotation of the members, a docking feature is connected to the docking port using the robot's crawling ability (Fig.~\ref{Fig_VTTDocking}~(d)).
A latch locks the connected member when the docking feature is fully inserted into the docking port.

\subsection{Experiment Design}

Planar robot configurations---e.g., configurations where all modules lay flat on the ground---cannot create out-of-plane motion merely by adding more modules since the Jacobian mapping the joint space to the 3D workspace is singular.
In other words, the robot actuators cannot exert a force in the vertical out-of-plane direction.
To solve this problem, the environment topology can be exploited to cause out-of-plane deformation in the 2D robot topology and thereby move the system out of the singularity. 

Here, we introduce three environments that allow a planar robot to escape its singular configuration. 
The first two environments enable a planar robot to self-reconfigure to a 3D topology. 
The skylight environment allows one robot to assist another in its 2D to 3D transformation.  The newly formed 3D robot can then crawl away---making way for another robot to undergo the same process.

\subsubsection{Mound Environment.}

\begin{figure*}[ht]
    \centering
    \includegraphics{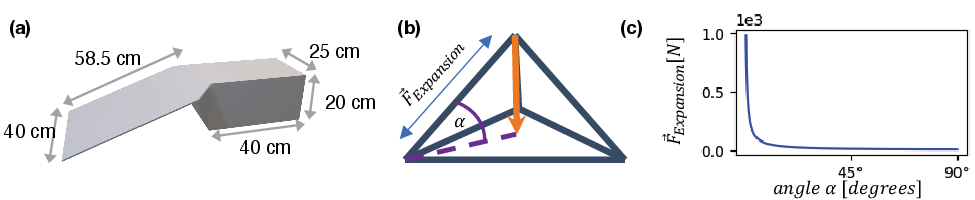}
    \caption{Pop-up tetrahedron mound environment and the relationship between elevation angle and expansion force: (a) mound feature dimensions; (b) tetrahedron diagram specifying the downforce due to gravity (orange arrow), the elevation angle $\alpha$, and the expansion force $\bm{F}_{\textit{Expansion}}$ of a single upper link; (c) graph highlighting the non-linear relation between elevation angle $\alpha$ and the total expansion force ($\bm{F}_{\textit{Expansion}}$) required to lift the top vertex up. 
    }
    \label{Fig_MoundEnvironment}
\end{figure*}

For this experiment, we consider a mound as a raised surface with a smooth slope. 
The size and type of mound that can be exploited depends on the size and type of robot platform. 
Fig.~\ref{Fig_RealVsArtificialEnv}(a) shows a profile sketch of the experiment environment used, a picture of the experiment setup, and a picture of mounds at the Mima Mounds Natural Preserve in Washington state.

The mound environment used in our experiment comprises a \SI{20}{cm} tall plywood box with a \SI{58.5}{cm} long ramp placed on a level surface (see Fig.~\ref{Fig_MoundEnvironment}(a)). 
The critical feature exploited in this experiment is the sloped surface of the ramp that allows the robot to elevate itself by crawling on top of the box. 
The box elevates the central vertex of the tetrahedron (see Fig.~\ref{Fig_MoundEnvironment}(b)) out of the singularity. 
The highly nonlinear, inverse relationship between the elevation angle and the force required to raise the tetrahedron structure means that even a small increase in elevation angle reduces the force required to pop up the tetrahedron by multiple orders of magnitude (see Fig.~\ref{Fig_MoundEnvironment}(c)).
\begin{equation}
    \bm{F}_{\textit{TotalExpansion}} = \frac{
    \textit{mass} \times g}{\sin \alpha},
\end{equation}
The above function (plotted in Fig.~\ref{Fig_MoundEnvironment}(c)) can be derived from the geometry of the links in a free body diagram.

This equation allows us to calculate if a tetrahedron flat-pattern configuration has crawled over a tall enough mound so the actuators can supply the required expansion force without  breaking its connections.

\subsubsection{Step Environment.}

The step environment features two surfaces separated by a \SI{30}{cm} tall step. 
A cylindrical obstacle is located on the lower surface approximately \SI{25}{cm} after the step. 
See Fig.~\ref{Fig_RealVsArtificialEnv}(b) for a profile sketch of the experiment environment, a picture of the experiment setup, and a picture of close-set slanted rocks on a hill in a forest that mimic a similar geometrical arrangement. 
This type of environment is not uncommon in urban environments: sloped steps and shallow banks with a drop are commonly found in architectural features. 
Similarly, natural environments provide plenty of inspiration: angled slate slabs with vertical drops, rock formations in forests, and eroded sandstone topologies in desert environments.

The experiment environment was constructed from \SI{25}{mm}~$\times$~\SI{25}{mm} aluminum extrusions and plywood. 
The surface was covered with a low-pile carpet to provide a consistent surface texture. 
The obstacle is a vinyl-wrapped \SI{25}{cm} cardboard tube with a heavy object inside to stabilize it.

\subsubsection{Skylight Environment.}

Instead of an obstacle, this environment features a foam-board ceiling with a teardrop-shaped cutout in it. 
The foam-board ceiling is level, while the carpeted floor beneath it is sloped downwards. 
This construction mimics a cave with a skylight opening in the ceiling or and overhanging canyon that leaves a narrow gap at the top (see Fig.~\ref{Fig_RealVsArtificialEnv}(c)).
This design was inspired by the Bowtie Arch and Antelope Canyon in Utah, and Gunsight Cave in Lincoln National Forrest in New Mexico.

This experiment environment uses the same construction and surface carpeting of the step environment. 
The distance between the ceiling and the slanted surface below is approximately \SI{24}{cm} at the center of the ceiling platform. 
The experiment was filmed from multiple angles to provide non-obstructed view angles at each stage. 



\subsection{VTT Simulation}

\begin{figure*}
    \centering
    \includegraphics[width=6.6in]{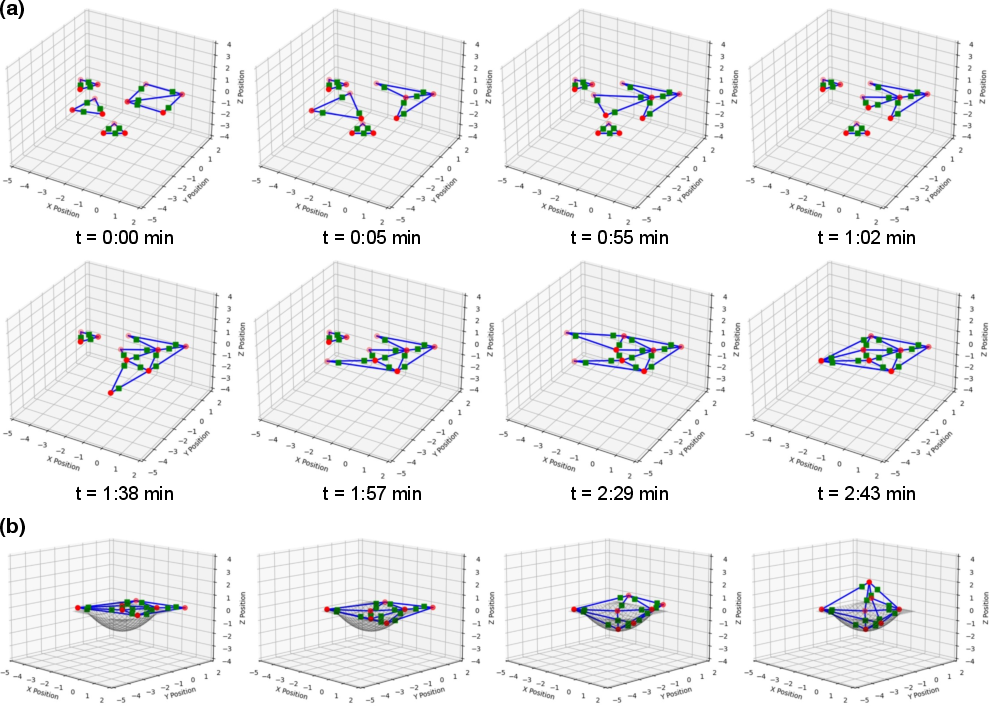}
    \caption{VTT Crawling reconfiguration simulation : (a) A set of three triangle-shaped VTT units navigates toward the target node and docks with the primary VTT structure (b) Transition from a 2D to 3D tetrahedral configuration is achieved through the pit-induced environmental condition}
    \label{Fig_VTTCreepExperimentSimulated}
\end{figure*}

In the simulation, the ground contact points of each VTT member (i.e., its sliding pad position) is set \SI{0.54}{m} away from the connector node and fixed to the actuator housing (see Fig.~\ref{Fig_VTT2DTo3D}(a) and (b)). Thus, the sliding pad distance from the actuator-side node remains constant even when the linear zipper expands.
Based on the approximate mass distribution, 65\% of the total mass is on one end of the spiral zipper and 35\% to the other end, with a total mass of \SI{1.92}{kg}. 
Assuming Coulomb friction, the friction coefficients were set using experimentally measured values --- static friction coefficient is 0.55 while the kinetic friction coefficient is 0.43.

The movement of the VTT is calculated by quasi-static analysis.
After the friction force is derived, the node velocities are calculated by combining the node's mass and damping. 
The node positions are updated by integrating the updated velocity. Normal forces are calculated based on the position of the center of mass. Then the accelerations of the nodes are calculated from the forces and integrated to compute the change in velocities. 

 Fig.~\ref{Fig_VTTCreepExperimentSimulated} shows the simulated process for 14 members starting as 4 separate planar robots that join to form a flattened octahedron, then transition to a 3D octahedron.

\section{Results}

In our experiments, the robots exploit their environment to reconfigure in a way that would not have been possible without making use of those environmental features. 
In the first scenario, a tetrahedron was formed from a 2D flat pattern within the simulation. In subsequent scenarios, each step was tested in real experiments, including 2D crawling to a target location, self-docking, and the transition from 2D to 3D configuration.
In this section, we include from video-frame sequences that show the robot reconfiguration process for each studied environment and analyze the challenges involved.

\subsection{VTT Crawling and Reconfiguration Simulation}

First, we extend each member at a speed of \SI{0.02}{m}/{s} towards the target length. Using the crawling motion, three triangle-shaped VTT units move toward the target node and dock with the main VTT robot. Once all individual robots are connected, we control the system to reconfigure into a tetrahedron flat-pattern. Then, exploiting a pit in the environment plane,  the central node moves out of plane by dropping into the pit. This motion breaks the flat-pattern from its singular position, and prepares it to rise to its 3D tetrahedron form using its motors. By adjusting the length of each member, the tetrahedral structure transitions from a 2D to a full 3D configuration.

\subsection{VTT Crawling Locomotion}

\begin{figure*}
    \centering
    \includegraphics[width=6.5in]{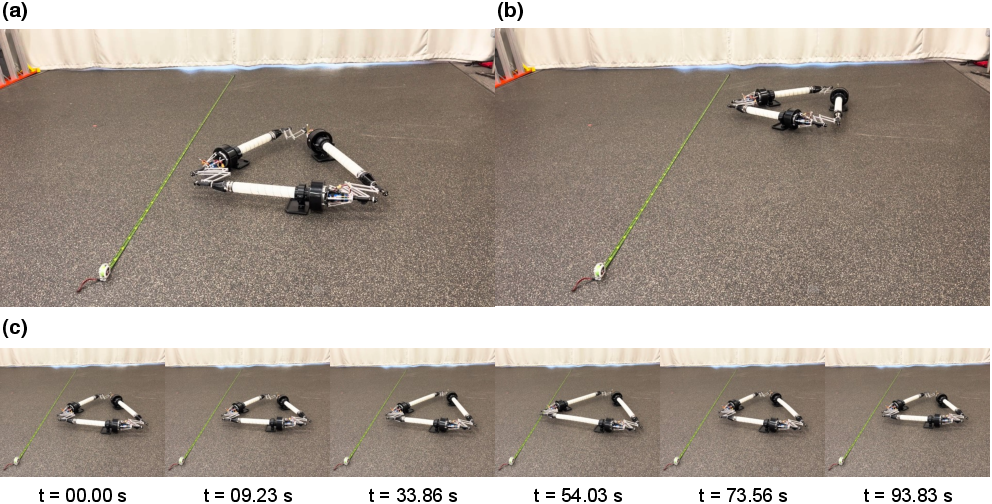}
    \caption{VTT crawling experiment: (a) Initial position of the VTT; (b) The final position of the VTT after 20  steps; (c) Frame sequence showing one cycle of VTT crawling locomotion.}
    \label{Fig_VTTCreepExperiment}
\end{figure*}

We tested the crawling locomotion for a 2D-triangular configuration on a flat surface.
The VTT was controlled by a pre-planned, open-loop controller that executed the crawling motion. The center-of-mass was shifted to change the friction force of each contact point as needed to execute the crawling behavior (see Fig. \ref{Fig_VTTCreepLocomotion}).
The experiment was conducted five times, and we confirmed that we can make the VTT move towards the desired direction with crawling locomotion strategy.
The average distance of travel per crawling cycle is \SI{8.3}{cm}, while the average drift in the perpendicular direction is \SI{1.5}{cm}.

\subsection{Self-Reconfiguration on a Flat Terrain}

\begin{figure*}
    \centering
    \includegraphics[width=6.6in]{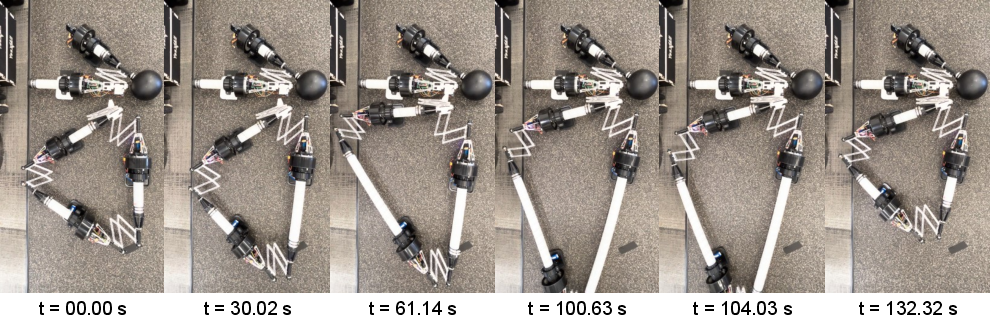}
    \caption{Snapshots of VTT docking. From t~=~0:00~min to t~=~4:53~min,  one triangular VTT crawls towards a target docking port of another triangular VTT. At t~=~4:53~min, the two VTT configurations lock with the locking lever.}
    \label{Fig_VTTDockingExperiment}
\end{figure*}

This experiment tested whether a triangle VTT configuration could successfully attach to another free connector using only its own locomotion abilities.
Fig.~\ref{Fig_VTTDockingExperiment} shows the docking process with the free VTT connector using only its own crawling locomotion.
In the experiment, the triangular VTT successfully approaches the docking port.
Once this motion inserts the dock  into the docking port, a servo moves a locking lever and the connection is secured.

\subsection{Self-Reconfiguration Using a Mound}

\begin{figure*}
    \centering
    \includegraphics{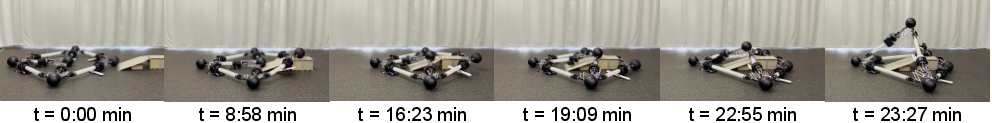}
    \caption{2D to 3D self-reconfiguration using the mound environment. The frame sequence shows the 2D to 3D transformation of  VTT. The small box with a ramp mimics a mound and allows the tetrahedron flat-pattern to elevate itself, such that the force required to raise the top three links of the tetrahedron are within the force capability of  VTT’s actuators.}
    \label{Fig_VTT2DTo3D}
\end{figure*}

In this experiment, the VTT crawls as a flattened tetrahedron pattern comprising seven links over  a mound obstacle and transforms into a tetrahedron. 

In the original VTT design the rotating zipper actuator drum would contact the ground and negatively affect the VTTs crawling behavior.
This happened because both the connector sphere and the spiral zipper are the same  \SI{20.3}{cm} (i.e., 8-inch) in  diameter.
To address this issue, We added FDM-printed sliding feet to the VTT (see Fig.~\ref{Fig_VTT2DTo3D}(a) and (b)).

To position itself on top of the mound, the tetrahedron flat pattern shaped robot executes a pre-programmed gate that first moves its central vertex forward, then  the two middle vertices on each side, then the trailing vertex. Finally, it extends the front vertex to ready itself for the next crawl cycle. 

Fig.~\ref{Fig_VTT2DTo3D} shows the successful transistion from 2D to 3D configuration. One challing aspect of this maneuver is when the central node must crest the edge of the box and makes contact with the linkage mechanism (see Fig.~\ref{Fig_VTT2DTo3D} t~=~16:23~min).

\subsection{Tetrahedron Formation Using the Step Environment}

\begin{figure*}
    \centering
    \includegraphics[width=6.6in]{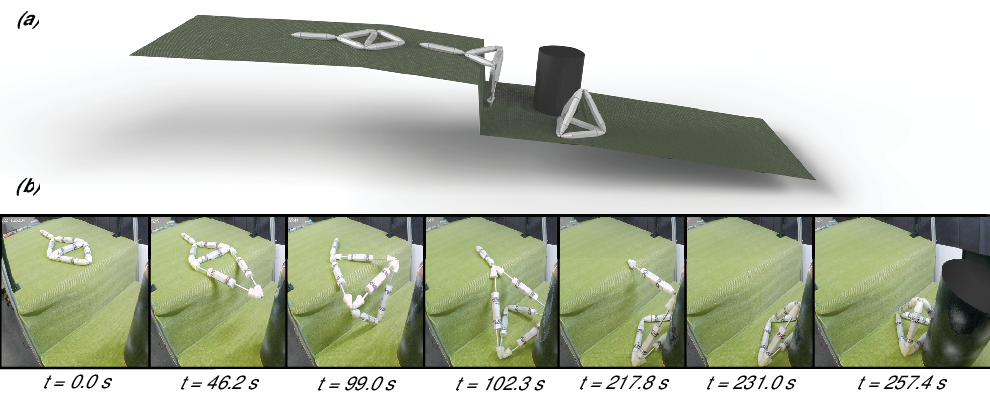}
    \caption{Self-reconfiguration using the step environment: (a) a 3D rendering of the self-reconfiguration process. (b) A frame sequence of the physical Truss Link flat pattern folding itself into a tetrahedron topology by exploiting the step and obstacle in the environment.}
    \label{Fig_TetStepEnv}
\end{figure*}

The transformation starts from a tetrahedron flat-pattern: the six-link diamond-with-tail topology shown on the left in Fig.~\ref{Fig_TetStepEnv}(a). 
Starting on the upper surface before the step, the robot shuffles over the ledge until the front half of the diamond folds downward (see Fig.~\ref{Fig_TetStepEnv}(b) t~=~0.0~s to t~=~99.0~s). 
Once the tip of the diamond touches the lower-level surface, the remaining links continue shuffling downwards until the diamond-cross-bar link touches the obstacle and slides to the lower surface—folding in such a way that only the tail link maintains contact with the top surface (see Fig.~\ref{Fig_TetStepEnv}(b) t~=~102.3~s to t~=~217.8~s). 
Finally, the robot retracts the tail link until it falls off the step and connects at the front vertex of the diamond-with-tail shape, as shown at t~=~231.0~s in Fig.~\ref{Fig_TetStepEnv}. 

The environment topology is crucial to enable this transition. 
The shallow slopes serve two purposes: first, they allow the robot move more easily by shuffling itself into position; second, they help the robot lean against the obstacle while pulling the tail link off the step and onto the free-vertex to form the tetrahedron. 
The obstacle stabilizes the under-constrained diamond-with-tail topology during the dynamic transition from the step to the lower-level surface. 
Without the obstacle, the diamond-with-tail would be significantly less likely to successfully transition to a tetrahedron topology since the tail link is outweighed by the diamond links.
The single link would move uncontrollably once the pull of the falling diamond links overcomes the single-link’s friction on the top surface.

It took the manual operator a total of 43 attempts to achieve the tetrahedron formation. This includes attempts for adjusting the environment setup, and include the learning curve by the operator for controlling a rather unintuitive machine with noticeable controller delay. The environment was adjusted five times over the course of these attempts. The successful attempt reported here was achieved in the sixth attempt after the last adjustment was made to the environment. This experiment was initially published in \cite{wyder2024robotlinks} and is further analyzed here to shed light on the use of environment exploitation.

\subsection{Assisted Reconfiguration Using the Skylight Environment}

\begin{figure*}
    \centering
    \includegraphics[width=6.6in]{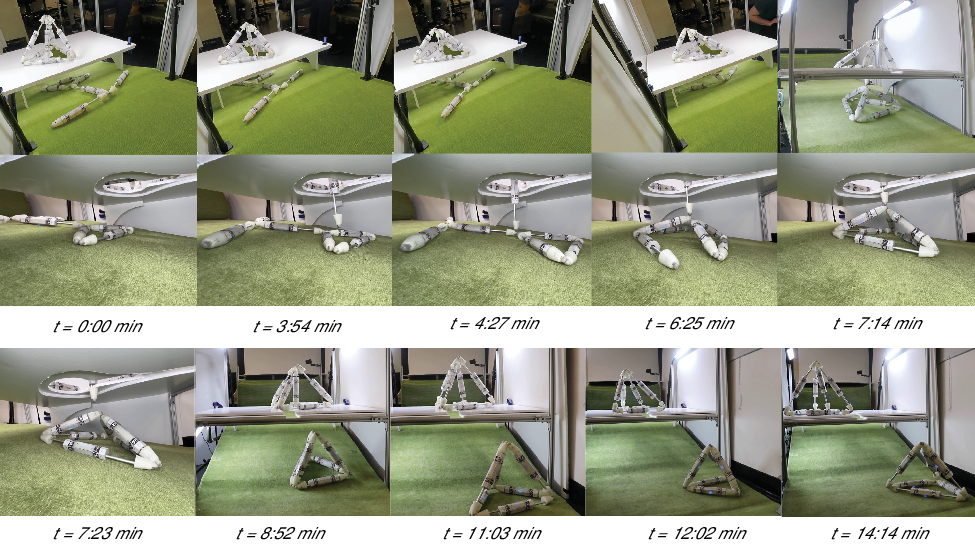}
    \caption{Assisted reconfiguration using the skylight environment: The ratchet tetrahedron positions itself above the skylight, and fishes with its dangling ratchet-link for the central vertex of the flattened tetrahedron pattern. Once connected, the ratchet tetrahedron lifts the central vertex up, and secures itself on the edge of the skylight opening until the Truss Links below have rearranged themselves into a tetrahedron robot by forming the base triangle (see t~=~4:27~min to t~=~7:14~min). Once completed, at t~=~7:23~min, the ratchet tetrahedron detaches from the newly formed tetrahedron. At this point the ratchet tetrahedron would be ready to assist the next flattened tetrahedron pattern or if finished could simply crawl away.}
    \label{Fig_TetAssist}
\end{figure*}

Robots can exploit their favorable position in an environment topology not only to transform themselves, but also to assist other robots in their transformations. 
The skylight environment has two levels: a flat roof level, and a sloped ground level that is accessible through a skylight opening in the roof. 
This experiment assumes that a three-dimensional topology has already been formed---a ratchet tetrahedron, i.e., a tetrahedron with an extra link attached to its top vertex. 
The ratchet tetrahedron then uses the extra link to lift the flattened tetrahedron pattern up and thereby enables it to form a tetrahedron. 
See Fig.~\ref{Fig_TetAssist} for a frame sequence detailing this transformation.
A key realization from these experiments is that environment exploitation can have wide-ranging implications. 
Helper robots such as this can enable the construction of many tetrahedron robots in a more repeatable manner. An environment feature may thereby become repeatedly exploitable, similarly to a spider positioning its web downwind to repeadetly catch pray.

\section{Limitations and Future Work}

A primary limitation of this study lies in the contrived nature of the experiment environments. While each scenario was designed to resemble plausible, commonly occurring  natural and architectural settings, the setups were purpose-built to highlight specific environment-robot interactions. The transferability of these behaviors to truly unstructured, real-world environments goes beyond the scope of this paper and is left for future work. Transferring these behaviors requires that the robot is capable of dealing with environment variations, and has the ability to recognize, localize, and then adaptively exploit them.

Additionally, the range of scales and platforms examined was limited. Features like the mound exploited by the VTT would not be usable in the sam way by smaller-scale systems such as Truss Links, and vice versa. This underscores that environment exploitation is inherently scale- and platform-dependent.

A broader challenge is the sheer vastness of the design space. A robot’s ability to exploit its surroundings depends on many factors: its geometry, actuation constraints, connection mechanics, and the structure of the environment. This high-dimensional, combinatorial search space is difficult to navigate using analytic methods alone.

In this context, manual experimentation is a practical and necessary first step.  It enables the isolation of key affordances and provides empirical grounding for studying environment-robot co-adaptation. But, due to the brittle nature of open-loop control methods on complex systems such as the Truss Links, this type of experimentation is impractical. Therefore, future progress will require simulation-based exploration at scale. In particular, we plan to generate families of synthetic environments that intentionally contain topological bottlenecks—scenarios that cannot be solved without leveraging environmental features. In such cases, robots would need to learn how to spot and exploit these features to self-reconfigure or self-assemble. This can be approached through structured search, evolutionary methods, or reinforcement learning.

A natural extension is the development of a taxonomy of environment exploitation strategies, such as elevation assistance, constraint-induced folding, or cooperative lifting. These can help generalize findings across platforms and guide design. Benchmarking metrics will also be needed—quantitative measures that assess how effectively a robot identifies, evaluates, and capitalizes on environmental features. These metrics should reflect not just success rates, but energy efficiency, robustness to uncertainty, and generalization across settings.

For this to work, robots must be able to perceive and interpret relevant environmental cues—such as slope, friction, accessibility, or constraint geometry—and select appropriate behaviors to exploit them. This calls for integrated perception-action loops tuned to environment morphology.

Finally, while this work focused on terrestrial environments, other domains---underwater, aerial, or space---offer radically different physical constraints and new opportunities for exploitation. For instance, buoyancy and drag in aquatic settings or microgravity in orbit could enable entirely novel reconfiguration behaviors. Extending these ideas beyond ground-based scenarios is a rich direction for future research.

\section{Discussion}

In this work, we explored how truss modular robots can actively exploit their environment to self-reconfigure and self-assemble. Across four experimental scenarios, we showed how environmental features such as elevation changes, ledges, and elevated platforms can compensate for robot limitations---particularly the inability to reconfigure from a planar configuration to a 3D topology.

We show that a shift in emphasis, from avoiding environmental features to exploiting them, allows us to make robots more capable. Rather than adding additional degrees of freedom to the robot modules, robots compensate by intelligently using their environment. This echoes strategies found in nature: organisms regularly manipulate or exploit environmental structure to support movement, construction, or adaptation. In robotics, this principle remains underexplored.

The environments in our experiments were deliberately structured to reveal specific forms of environment-robot synergy. While this limits generalization, it provides necessary groundwork for future automated discovery and design of such interactions.

Ultimately, we believe that exploiting the environment should be seen not as a fallback, but as a core competency for robots in general. As we move toward practical deployment, intelligent environmental interaction may prove essential for reducing complexity, increasing robustness, and scaling to real-world applications.

\section{Funding Statement}
The author(s) disclosed receipt of the following financial support for the research, authorship, and/or publication of this article: This work was partially funded by DARPA TRADES COLUM 5216104 SPONS
GG012620 01 60908 HL2891 20 250, NSF NRI COLUM 5216104 SPONS
GG015647 02 60908 HL2891, and NSF NIAIR COLUM 5260404 SPONS
GG017178 01 60908 HL2891.

\bibliographystyle{unsrt}  

\bibliography{bibliography}

\end{document}